\documentclass{bmvc2k}
\usepackage{adjustbox}
\usepackage{booktabs}
\usepackage{multirow}
\usepackage{multicol}
\usepackage{amssymb}
\usepackage{pifont}
\newcommand{\cmark}{\ding{51}}%
\newcommand{\xmark}{\ding{55}}%

\usepackage{wrapfig}
\DeclareUnicodeCharacter{2212}{-}

\usepackage{floatrow}
\newfloatcommand{capbtabbox}{table}[][\FBwidth]

\bmvcreviewcopy{1497}

\title{MMD-ReID: A Simple but Effective solution for Visible-Thermal Person ReID}

\addauthor{Chaitra Jambigi}{http://www.vision.inst.ac.uk/~ss}{1}
\addauthor{Ruchit Rawal}{http://www.vision.inst.ac.uk/~pp}{1}
\addauthor{Anirban Chakraborty}{colin@collaborators.com}{1}

\addinstitution{
 The Vision Institute\\
 University of Borsetshire\\
 Wimbleham, UK
}
\addinstitution{
 Collaborators, Inc.\\
 123 Park Avenue,\\
 New York, USA
}

\runninghead{Student, Prof, Collaborator}{BMVC Author Guidelines}

\def\etal{\emph{et al}\bmvaOneDot}

\begin{document}

\maketitle

\begin{abstract}

Learning modality invariant features is central to the problem of Visible-Thermal cross-modal Person Reidentification (VT-ReID), where query and gallery images come from different modalities. Existing works implicitly align the modalities in pixel and feature spaces by either using an adversarial learning framework or carefully designing feature extraction modules that heavily rely on domain knowledge. In this work, we propose a simple but effective framework, MMD-ReID, that reduces the modality gap by an explicit discrepancy reduction constraint. MMD-ReID takes inspiration from Maximum Mean Discrepancy (MMD), a widely used statistical tool for hypothesis testing that determines the distance between two distributions. MMD-ReID uses a novel margin-based formulation to match class conditional feature distributions of visible and thermal samples to minimize intra-class distances while maintaining feature discriminability. MMD-ReID is a simple framework in terms of architecture as well as loss formulation. We conduct extensive experiments to demonstrate both qualitatively and quantitatively the effectiveness of MMD-ReID in aligning the marginal and class conditional distributions, thus learning both modality-independent and identity-consistent features. The proposed framework significantly outperforms the state-of-the-art methods on SYSU-MM01 and RegDB datasets.

\end{abstract}

\section{Introduction}
\label{sec:intro}


Person re-identification (ReID) has been extensively studied in the computer vision literature as a pedestrian matching problem between query and gallery images from different cameras \cite{ye2021deep, Wang2014PersonRB, WangMatrix}. Traditionally, most methods focus on scenarios where single-modality visible cameras capture images, i.e., visible-visible ReID (VV-ReID), wherein the focus is on matching visible images. However, in 24-hour intelligent surveillance systems, we need to process data from infrared cameras at nighttime. Consequently, there has been a significant interest in Visible-Thermal ReID (VT-ReID) which, given a visible image, aims to match it to the thermal image of the same person \cite{wu2017rgb, ye2018visible, Dai2018CrossModalityPR, wang2019rgb}. VT-ReID is more challenging than VV-ReID as it suffers from both intra-modality variations (caused by pose, illumination, and viewpoint changes) as well as inter-modality variations (caused by a huge modality gap between visible and thermal images \cite{WangDualDisc, wang2019rgb, lu2020cross, Hao_Wang_Li_Gao_2019}).

The quest to bridge the cross-modality discrepancy has pushed advancements in two significant directions: First, adversarial-learning based approaches have paved the way for joint pixel and feature space alignment \cite{wang2019rgb, Dai2018CrossModalityPR, Kniaz2018ThermalGANMC, WangDualDisc}. This is typically achieved by leveraging generative adversarial networks to translate an image from a heterogenous modality to the desired modality and by using a mini-max setup to learn modality invariant feature representations. However, generative methods do not guarantee identity preservation across modality translation and often require excessive training tricks and additional computation. Second, shared feature learning techniques currently achieve state-of-the-art results for VT-ReID by projecting features from heterogeneous modalities into a common feature space \cite{lu2020cross, liu2020parameter, ye_modality_collab, choi2020hi}. However, they heavily rely on carefully designed feature selection modules such as partition strips \cite{sunPCB, liu2020parameter, Ye2020DynamicDA}, pre-trained pose-driven matching \cite{xu_pose, su_pose}, semantic alignment \cite{Kalayeh2018HumanSP}, human landmarks \cite{Wang2020HighOrderIM} to discard modality-specific redundant information. In light of these observations, recent studies \cite{Luo2019BagOT, liu2021strong, Chen_2021_CVPR} have criticized the current state-of-the-art methods' overly complex and rigid nature, citing the need for new algorithmic ideas that are both simple and effective.

We approach the problem of learning modality invariant representations in the VT-ReID task from an explicit distribution discrepancy perspective. The centerpiece of such a formulation is the use of a statistical hypothesis testing framework called maximum mean discrepancy (MMD) \cite{gretton2012kernel} that measures the proximity between two distributions. MMD has been widely studied in unsupervised domain adaptation (UDA) literature to minimize marginal \cite{long_dan, Long2016UnsupervisedDA} as well as (more recently the) class-conditional distribution discrepancy \cite{Yan2017MindTC}. Inspired by this, we adopt MMD in the supervised VT-ReID task to align visible and infrared distributions for a particular identity. However, we (empirically) observed that this formulation is vulnerable to overfitting and feature degradation, leading to suboptimal results. To alleviate this problem, we introduce a novel margin-based MMD loss: Margin MMD-ID.

With the goal of providing a simple yet strong framework to achieve competitive performances, we propose MMD-ReID that utilizes Margin MMD-ID as its core training objective. MMD-ReID is simple, primarily as (1) it only uses the global features and does not rely on part-level features. (2) It is easily extendable since it's built on the traditional two-stream network that has enjoyed promising results in VT-ReID. Furthermore, (3) Margin MMD-ID loss is intuitive and easy to train in a deep learning setup. We demonstrate the effectiveness of MMD-ReID through extensive experimentations on two popular benchmark datasets: SYSU-MM01 and RegDB, outperforming the current state-of-the-art by 5.07\% and 4\% Rank1 accuracy, respectively. Moreover, we empirically observe that our modified loss function: Margin MMD-ID is not only complementary to the current best practices in the ReID community but can also be easily adopted in existing baselines to boost the performance further. 

In summary, the main contributions of our work are:
\begin{itemize}
    
    \item We propose a simple but effective framework: MMD-ReID, which to the best of our knowledge, is the first work to explore the VT-ReID task from the perspective of explicit distribution discrepancy reduction constraint. MMD-ReID employs our novel margin-based modification: Margin MMD-ID loss to alleviate the problem of overfitting and feature degradation that occurs with standard MMD in supervised VT-ReID.
    
    
    \item Extensive experiments demonstrate that MMD-ReID achieves state-of-the-art results on two benchmark datasets: SYSU-MM01 and RegDB. It is worth mentioning that we achieve improvement in performance just by using global features.
    
    \item We empirically demonstrate that Margin MMD-ID can be used on top of existing baselines to improve their performance further. We verify our claim by performing experiments on three popular baselines: AGW \cite{ye2021deep}, Hc-Tri \cite{liu2020parameter}, DGTL \cite{liu2021strong}.
    
\end{itemize}

\section{Related works}
\label{sec:related}
    \textbf{VV-ReID}: Person re-identification problem has been primarily studied in a closed-world setting where images are acquired by single-modality cameras \cite{martinel_deep, zheng_bench, Zhang2017AlignedReIDSH, Zhu2020ViewpointAwareLW}. Accordingly, researchers have focussed a great deal of attention on dealing with challenges of appearance changes pertaining to a single modality, such as variation in viewpoint \cite{karnam_viewpoint, bak_view}, pose \cite{zhao_pose, Sarfraz2018APE, cho_pose}, illumination \cite{huang_illum}, occlusions \cite{Hou2019VRSTCOV}, and background clutter \cite{song_clutter}. This is usually achieved by augmenting standard convolutional neural networks with powerful (manually-designed) feature selection modules such as partition-strips \cite{sunPCB, liu2020parameter}, pose estimation \cite{Zheng2019PoseInvariantEF} to handle occlusions and misalignment, etc. Another line of approaches utilizes deep metric learning \cite{liu2020parameter, Varior2016ASL, Deng2018ImageImageDA, deep_metric, Chen2017BeyondTL}  to design loss functions (such as triplet loss \cite{Hermans2017InDO}, quadruplet loss \cite{Chen2017BeyondTL}) that ensure robust and discriminative feature representations. 
    
    \textbf{VT-ReID}: Recent VT-ReID methods primarily rely on either adversarial learning-based modality alignment or modality shared feature learning methods to alleviate cross-modality discrepancy. Inspired by the success of generative adversarial networks, adversarial learning-based approaches aim to perform cross-modality alignment in pixel and feature space. Dai \etal \cite{Dai2018CrossModalityPR} utilized adversarial training strategies to learn modality invariant feature representations. Kniaz \etal.\cite{Kniaz2018ThermalGANMC} proposed a novel GAN framework ThermalGAN to translate a single visible probe image to a thermal probe set and perform conventional ReID in the thermal domain. Wang \etal \cite{wang2019rgb} in their work, employed an end-to-end three-player mini-max setup to jointly optimize for pixel and feature space alignment across modalities. On similar lines, Wang \etal \cite{WangDualDisc} proposed to decompose modality and appearance discrepancy and reduce them separately using a bi-directional cycleGAN and conventional feature level constraints, respectively. 
    
    Learning robust and discriminative shared feature representations is central to the success of VT-ReID systems. Most recent studies approach this through a two-stream network backbone (first proposed by Ye \etal \cite{Ye2018HierarchicalDL, ye2018visible, ye_bi}) that projects cross-modality embeddings in a common feature space. Ye \etal \cite{ye_modality_collab} in their work handle the modality discrepancy at both feature and classifier level by proposing an ensemble learning scheme to incorporate the modality shareable classifier and the modality-specific classifiers. Liu \etal \cite{LIU202011} in pursuit of learning robust and discriminative person features, proposed a mid-level feature incorporation strategy using skip-connections. Shared features disregard modality-specific features reducing the discriminability of feature representation. To alleviate this problem, Lu \etal \cite{lu2020cross} proposed a novel shared-specific feature transform algorithm to utilize both modality-specific and modality-shared information by modeling the affinities between intra-modality and inter-modality samples. Liu \etal \cite{liu2020parameter} in their work, proposed the hetero-center based triplet loss to provide a strong baseline for VT-ReID tasks utilizing both global and local feature extraction strategies. 

    \textbf{MMD}: In the scope of deep learning, MMD was first studied in the unsupervised domain adaptation (UDA) literature to align source and target distributions. Most notably, Long \etal.~\cite{long_dan} first introduced the idea of minimizing multi-kernel MMD between task-specific layers to enhance feature transferability across domains. To optimize conditional-distributions discrepancy, Long \etal \cite{long_pseudo} adopted a pseudo label refinement strategy to generate target domain labels and perform a joint adaptation of both marginal and conditional distributions between domains. Owing to its intuitive and strong foundations, MMD has been adopted by diverse emerging paradigms in deep learning such as generative adversarial networks, variational autoencoders, transfer-learning, noise-insensitive auto-encoders \cite{mmd_gan, Ragonesi2020LearningUR}.

\section{Methodology}
\label{sec:method}
The rest of the paper is organised as follows: Section \ref{MMD} briefly introduces MMD, Section \ref{MMD_in_VTReid} describes using MMD for VT-ReID task and margin-based modifications.  Section \ref{MMDReIDFramework} describes the architecture, batch sampling strategies and overall loss formulation. Section \ref{sec:res and analysis} describes in detail the datasets, experiments, results, and ablation studies. Section \ref{sec:conclusion} concludes the work with future directions.

\subsection{Maximum Mean Discrepancy (MMD)}
\label{MMD}
Let $P$ and $Q$ be two distributions and $\{X_n\}$,$\{Y_m\}$ be two datasets, $\{X_n\} \sim P $, $\{Y_m\} \sim Q $

The two sample test is one of the fundamental tests in statistics that tries to determine whether the given two datasets, $\{X_n\} \sim P $ and $\{Y_m\} \sim Q $ are generated from the same underlying distribution or not. This task is difficult since the distribution information is generally unknown apriori \cite{bickel1969distribution,biau2005asymptotic,hall2002permutation,friedman1979multivariate}. MMD is a test statistic that measures the discrepancy of two distributions by embedding them in a Reproducing Kernel Hilbert space (RKHS) \cite{gretton2012kernel}. To simplify, MMD performs the two sample test by finding the difference between the mean function values of the two samples evaluated on a smooth function, where the function class for MMD is a unit ball in an RKHS. If the difference in mean values is large, then the samples are likely to be drawn from different distributions. The formulation of MMD is,

{\footnotesize
\begin{align}
\begin{split}
\label{eq:1}
    {MMD}^{2}(X, Y) ={}& \left\|\frac{1}{N} \sum_{n=1}^{N} \phi\left(X_{n}\right)-\frac{1}{M} \sum_{m=1}^{M} \phi\left(Y_{m}\right)\right\|^{2}
\end{split}\\
\begin{split}
\label{eq:2}
      ={}& \frac{1}{N^{2}} \sum_{n=1}^{N} \sum_{n^{\prime}=1}^{N}  \phi\left(X_{n}\right)^{\top} \phi\left(X_{n^{\prime}}\right)+\frac{1}{M^{2}} \sum_{m=1}^{M} \sum_{m^{\prime}=1}^{M} \phi\left(Y_{m}\right)^{\top} \phi\left(Y_{m^{\prime}}\right)-\frac{2}{N M} \sum_{n=1}^{N} \sum_{m=1}^{M} \phi\left(X_{n}\right)^{\top} \phi\left(Y_{m}\right)
\end{split}\\
      ={}& \mathbf{E}_{x, x^{\prime} \sim P}\left[k\left(x, x^{\prime}\right)\right]+\mathbf{E}_{y, y^{\prime} \sim Q}\left[k\left(y, y^{\prime}\right)\right]-2 \mathbf{E}_{x \sim P, y \sim Q}[k(x, y)]
      \label{eq:3}
\end{align}
}%

where $\phi(.)$ is the feature mapping function.
\\
Kernel trick can then be applied on the inner product in Eq.\eqref{eq:2} to get Eq.\eqref{eq:3}

\subsection{MMD in VT-ReID}
\label{MMD_in_VTReid}
Let $P=\left\{{x}_{v}^{i}...{x}_{v}^{N_V}\right\}$ and $Q=\left\{{x}_{t}^{i}...{x}_{t}^{N_T}\right\}$ denote the visible and thermal images, respectively. \(N_V\) and \(N_T\) denote the total number of visible and thermal images in the dataset, respectively. To reduce the distribution discrepancy in the shared space, we use MMD distance as the criterion to explicitly learn representations such that the MMD loss between visible and thermal features is minimized. 
\begin{equation}\label{eq:4}
L_{MMD}(P,Q) =\underbrace{\mathbf{E}_{P}\left[k\left(x_v, x_v^{\prime}\right)\right]+\mathbf{E}_{Q}\left[k\left(x_t, x_t^{\prime}\right)\right]}_{same \;modality \;distribution}-\underbrace{2 \mathbf{E}_{P,Q}[k(x_v, x_t)]}_{cross \;modality \;distribution}
\end{equation}

The first two terms are the kernel similarity between the same modality samples, which has a high value at the start of training. The last term is the similarity between cross-modality samples, which is low initially. When MMD loss is minimized, it eventually tries to bring the cross-modality similarity as close as possible to the same modality similarity, thereby aligning both the distributions. MMD aims to match infinite order moments with a Gaussian kernel \cite{gretton2012kernel}. Thus, reducing the MMD distance aligns the two distributions in a superior way compared to other implicit aligning methods discussed in the introduction section (Section-\ref{sec:intro}).
\newline

\textbf{MMD-ID: }
The above MMD loss formulation in Eq.\eqref{eq:4} aligns the two modalities marginally without considering the class conditional distribution relationship between the two modalities. Thus, when modalities get aligned, the learned features may not preserve the class discriminative property. In order to align the modalities, respecting the class-wise distribution, we use a modified version of MMD, in which we precisely align the distributions on a per-identity basis and averaging over all possible identities. The modified loss is of the form,

\begin{equation}\label{eq:5}
MMD^{2}(P^c, Q^c) = \mathbf{E}_{P}\left[k\left(x_v^c, x_v^{c'}\right)\right]+\mathbf{E}_{Q}\left[k\left(x_t^c, x_t^{c'}\right)\right]-2 \mathbf{E}_{P,Q}[k(x_v^c, x_t^c)]
\end{equation}
\begin{equation}\label{eq:6}
L_{MMD-ID}(P,Q) =\frac{1}{C} \sum_{c=1}^{C} MMD^2(P^c, Q^c)
\end{equation}
$P^{c}$ and $Q^{c}$ denote visible and thermal sample distribution of a particular $c^{th}$ identity.
\newline

\begin{figure}[h]
    \centering
    \includegraphics[width=\textwidth]{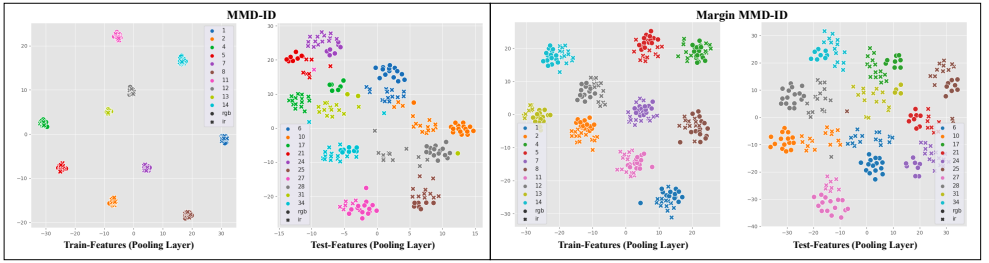}
    \caption{t-SNE plot for the last epoch of the model trained with MMD-ID and Margin MMD-ID. The intra-class compactness in train data doesn't translate to testing data indicating feature degradation and high overfitting.}
    \label{fig:mmd_overfit}
    \end{figure}

\textbf{Margin MMD-ID: }
Although MMD-ID is intuitive, it can suffer from the problem of overfitting, thus collapsing all the features of the same identity to a small region in feature space, as shown in Figure \ref{fig:mmd_overfit}. To mitigate this effect and optimally use the strengths of MMD-ID, we propose a new margin-based loss as,

\begin{equation}\label{eq:7}
MMD^{'2}(P^c, Q^c)= 
\begin{cases}
    MMD^2(P^c, Q^c),& \text{if } MMD^2(P^c, Q^c)- \rho > 0 \\
    0,              & \text{otherwise}
\end{cases}
\end{equation}
\vspace{-2mm}
\begin{equation}\label{eq:8}
L_{Margin-MMD-ID} =\frac{1}{C} \sum_{c=1}^{C} MMD^{'2}(P^c, Q^c)
\end{equation}
We add a margin term $\rho$, which can control the amount of distribution alignment, thus keeping a balance between aligned and generalised model. 
Intuitively, we measure the averaged MMD-ID distance over the training and restrict the reduction to a certain value, i.e. \(\rho\). 

\begin{figure}[t]
  \centering
  \includegraphics[width=0.95\textwidth, height=5cm]{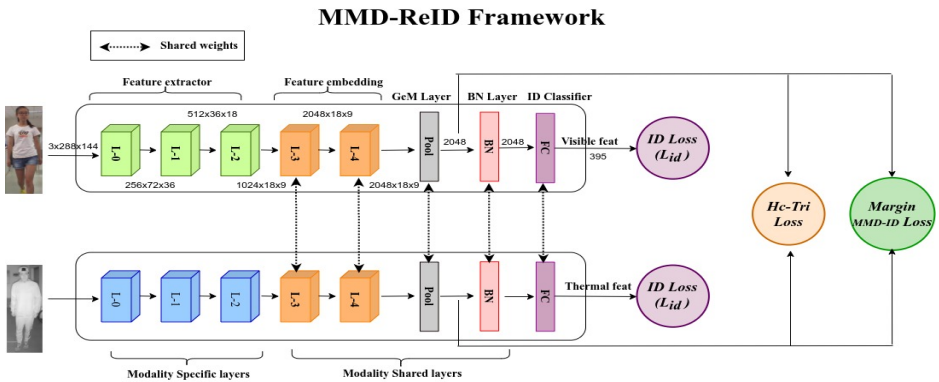}
  \caption{MMD-ReID: Structure of our two stream architecture for VT-ReID. Modality specific layers (L-0, L-1, L-2) have independent weights for each modality. Modality shared layers (L-3, L-4, Pool, BN, FC) have shared weights for both modalities, denoted by dotted by bi-directional arrows. Visible and Thermal features are extracted independently and ID loss is applied. Margin MMD-ID and Hc-Tri are applied on pooled features.}
  \label{fig:Figure 1}
\end{figure}

\subsection{MMD-ReID Framework}
\label{MMDReIDFramework}
We introduce our proposed framework MMD-ReID as depicted in Fig \ref{fig:Figure 1}. Our model mainly consists of two components: 1. Two stream backbone network to explore the shared and specific features  2. Our proposed Margin MMD-ID loss along with Identity softmax loss and triplet loss to get identity separable as well as discriminative features.
\newline
\textbf{Two stream network: }
We adopt the conventional two-stream architecture as \cite{liu2020parameter} which consists of feature extractor and feature embedding to extract modality-specific features and shared features, respectively. We use ResNet50 \cite{he2016deep} as the backbone with initial shallow layers and first two res-convolution blocks as feature extractor (L-0,L-1,L-2 in Fig. \ref{fig:Figure 1}) which have separate weights for each modality and last two res-convolution blocks (L-3,L-4) as feature embedding, followed by pooling and BN layers, which have shared weights for both modalities. To get fine-grained features, we use Generalized-mean (GeM) Pooling instead of average or max pooling \cite{radenovic2018fine, ye2021deep}. For details on GeM layer, refer supplementary material.
\newline
\textbf{Batch sampling: }
We create our mini-batch by randomly sampling $P\times K$ images, where $P$ is the number of identity in the batch and $K$ is the number of images per identity. We randomly choose $K$ visible and $K$ thermal images, per identity to mitigate class imbalance issues, effectively making a batch size of $2\times P\times K$.
\newline
\textbf{Overall loss: }
We use our proposed Margin MMD-ID loss Eq.\eqref{eq:8} along with the standard identity softmax loss to learn discriminative features. Our loss explicitly aligns the two modalities based on class conditional distributions, thereby reducing the intra-class discrepancy. However, inter-class separation is not guaranteed, which is needed for good representation learning in open-set problems. To tackle this, we use a variant of Triplet loss, called Hetero-center triplet loss (Hc-Tri) \cite{liu2020parameter} to maximize inter-class distances. Hc-Tri is formulated in the same way as standard Triplet loss \cite{hermans2017defense}, but it takes centers of different modalities as input rather than individual samples. More details on Hc-Tri are provided in the supplementary material. Although Hc-Tri also reduces the intra-class distances, it is worth mentioning that the space in which MMD and Triplet losses work is different. Triplet loss formulation brings anchor and positive closer in euclidean space, whereas the MMD loss statistically matches all the higher-order moments. Thus, MMD is a stronger loss in terms of distribution alignment as compared to Hc-Tri loss. The total loss is of the form,


\vspace{-2mm}
\begin{equation}\label{eq:10}
    L = \lambda_1 L_{id} + \lambda_2 L_{Margin-MMD-ID} + \lambda_3 L_{Hc-Tri}
\end{equation}

\vspace{-4mm}

\section{Experiments and Results}
\label{sec:res and analysis}
\subsection{Datasets and settings}

\textbf{SYSU-MM01:} SYSU-MM01 \cite{wu2017rgb} is a large-scale dataset containing images captured by two thermal and four visible cameras. It contains 491 identities and we use 395/96 identities for training/testing, making 22,258 visible and 11,909 thermal images for training. The test set contains 3803 thermal images for Query and 301 randomly selected visible images as Gallery. We adopt the most challenging and commonly used evaluation mode: All search/Indoor search in Singleshot setting, where only one gallery image per identity is available. We follow the evaluation protocol as \cite{liu2020parameter, ye2018visible, ye2021deep} to perform ten trials of gallery set selection and then report the average performance.

\textbf{RegDB:} The dataset \cite{nguyen2017person} is collected by dual-camera systems (visible and thermal) and includes 412 identities. For each identity, ten visible and ten thermal images are captured. We follow the evaluation protocol as \cite{choi2020hi, ye2018visible} where the dataset is randomly split into two parts, one for training and one for testing. For testing, images from one modality are selected as gallery and images from other modality as probe set. The process is repeated for ten trials and averaged results are reported.

\textbf{Evaluation metrics: }
Following standard protocol \cite{wu2017rgb}, Cumulative matching characteristics (CMC) and mean average precision (mAP) are adopted as evaluation metrics. Query and gallery are from different modalities. CMC (rank-k) measures whether correct identity from cross modality is retrieved in top-k results and mAP measures retrieval performance when the gallery set contains multiple matching images. 

\textbf{Implementation details:}
For implementation details please refer to the supplementary.


\subsection{Results and Analysis}
\paragraph{Comparison with state-of-the-art}

The results on SYSU-MM01 and RegDB datasets is shown in Table \ref{table:1}, \ref{table:2} respectively. All metrics for other methods are taken from their paper. In the All-search mode, our method surpasses the current state-of-the-art method: cm-SSFT by 5.15 \%, 4.96 \%, and 3.48 \% in rank-1, rank-10, and rank-20 metrics respectively while achieving comparable mAP. A similar trend is observed in the Indoor search where we significantly outperform the state-of-the-art on all metrics. We observe that we marginally lag behind ‘Farewell to Mutual Info.’ on rank-10 and rank-20 in All-search mode, however considerably surpass them in rank-1 and mAP as well as on all metrics in Indoor-search. Our results on RegDB are better than the state-of-the-art: Hc-Tri by 4\% on Rank-1 and by 5.67\% on mAP for Visible to Thermal task and the gain for Thermal to Visible task is of 4.35\% in Rank-1 and 5.84\% in mAP.

\begin{table}[ht]
\begin{center}
\begin{adjustbox}{width=0.7\linewidth}
\centering
\begin{tabular}{c|ccc|c|ccc|c}
\hline
\textbf{Method} & \multicolumn{4}{c|}{All Search} & \multicolumn{4}{c}{Indoor Search}\\
 \cline{2-9}
  &r1&r10 &r20&mAP &r1&r10 &r20&mAP\\
\hline
BDTR \cite{ye2018visible} & 17.01&55.43&71.96&19.66&-&-&-&-\\

SDL \cite{kansal2020sdl} & 28.12&70.23&83.67&29.01&32.56&80.45&90.67&39.56\\

cmPIG \cite{wang2020cross} & 38.1&80.7&89.9&36.9& 43.8& 86.2& 94.2& 52.9\\

Hi-CMD \cite{choi2020hi} & 34.94&77.58&-&35.94&-&-&-&-\\

CASE-Net \cite{li2020learning} & 42.9 & 85.7 & 94.0 & 41.5 & 44.1 & 87.3 & 93.7 & 53.2 \\

AlignGAN \cite{wang2019rgb} & 42.4 &85.0& 93.7& 40.7 & 45.9 &87.6 &94.4 &54.3\\

Neural Feature Search \cite{Chen_2021_CVPR} & 56.91& 91.34& 96.52& 55.45& 62.79& 96.53& 99.07& 69.79\\

Farewell to Mutual Info \cite{Tian_2021_CVPR} & 60.02& \textbf{94.18}& \textbf{98.14}& 58.80& 66.05 &96.59 &99.38 &72.98\\

Hc-Tri \cite{liu2020parameter} & 61.68 & 93.10& 97.17& 57.51& 63.41 &91.69 &95.28 &68.17 \\


cm-SSFT \cite{lu2020cross} & 61.6& 89.2& 93.9& \textbf{63.2}& 70.5 &94.9 &97.7 &72.6\\

MACE \cite{ye_modality_collab} & 51.64 & 87.25 & 94.44& 50.11& 57.35& 93.02& 97.47&64.79 \\

MMD-ReID (Ours) & \textbf{66.75}& 94.16 & 97.38 & 62.25 & \textbf{71.64}& \textbf{97.75}& \textbf{99.52}& \textbf{75.95}\\
\hline
\end{tabular}
\end{adjustbox}
\end{center}
\vspace{-5mm}
\caption{Results on SYSU-MM01 dataset}
\label{table:1}
\end{table}


\begin{table}[ht]
\begin{center}
\begin{adjustbox}{width=0.7\linewidth}
\centering
\begin{tabular}{c|ccc|c|ccc|c}
\hline
 \textbf{Method} & \multicolumn{4}{c|}{\textbf{Visible to Thermal}} & \multicolumn{4}{c}{\textbf{Thermal to Visible}}\\
 \cline{2-9}
  &r1&r10 &r20&mAP &r1&r10 &r20&mAP\\
\hline

BDTR \cite{ye2018visible} & 33.47 & 58.42 & 67.52 & 31.83 & 32.72 &   57.96 &  68.86 & 31.10\\

SDL \cite{kansal2020sdl}  &26.47&51.34&61.22&23.58&25.74&50.23&59.66&22.89\\

cmPIG \cite{wang2020cross}  &48.5&-&-&49.3&48.1&-&-&48.9\\

Hi-CMD \cite{choi2020hi} &70.93&86.39&-&66.04&-&-&-&-\\

AlignGAN \cite{wang2019rgb} &57.9&-&-&53.6&56.3&-&-&53.4\\

Neural Feature Search \cite{Chen_2021_CVPR} &80.54 &91.96 &95.07 &72.10&77.95 &90.45 &93.62 &69.79\\

Farewell to Mutual Info \cite{Tian_2021_CVPR} &73.2&-&-&71.6&71.8&-&-&70.1\\

Hc-Tri \cite{liu2020parameter} &91.05 &97.16 &98.57 &83.28& 89.30 &96.41 &98.16 &81.46\\


cm-SSFT \cite{lu2020cross} &72.3&-&-&72.9&71.0&-&-&71.7\\

MACE \cite{ye_modality_collab} & 72.37 & 88.40 & 93.59& 69.09& 72.12& 88.07& 93.07&68.57 \\

MMD-ReID (Ours) & \textbf{95.06} & \textbf{98.67} & \textbf{99.31} & \textbf{88.95} & \textbf{93.65} & \textbf{97.55} & \textbf{98.38} & \textbf{87.30}\\
\hline
\end{tabular}
\end{adjustbox}
\end{center}
\vspace{-5mm}
\caption{Results on RegDB dataset}
\label{table:2}
\end{table}

\vspace{-4mm}

\paragraph{Ablation study of different loss components:}

Table \ref{table:3} shows the importance of each loss component in training. It is evident that using only cross-entropy loss (CE), or CE with Hc-Tri (row:1,6) loss gives sub-optimal results, and thus there is a scope for explicit modality alignment. We observe a boost in both rank-1 and mAP after adding MMD loss with CE (row 2) in both the datasets, which supports our claim that explicit discrepancy reduction helps in VT-ReID. We further see that replacing MMD with MMD-ID (row 3) rather drops the mAP and rank-1 by \(\sim\) 2\% for the SYSU-MM01 dataset, and the reason for this is the overfitting of the model leading to feature degradation as shown in Fig.\ref{fig:mmd_overfit}. To regularise this, we add a margin term in MMD-ID as per Eq.\eqref{eq:7} and we see an increase in rank-1 and mAP indicating a reduction in misclassifications (row 4) which is in agreement with Fig.\ref{fig:mmd_overfit}. Further adding Random erasing (RE) as augmentation helps in the overall generalization of our model giving the best accuracy in row 5. In a complementary sense, since Margin MMD-ID cannot increase inter-class distances, we adopt Hc-Tri loss for this purpose. As discussed in Section \ref{MMDReIDFramework}, although Hc-Tri loss reduces intra-class distances, MMD is a stronger loss in terms of distribution alignment, hence using Margin MMD-ID with Hc-Tri performs better than only Hc-Tri which can be shown from rows 6,9. Row 6-10 is similar to Row 1-5 but with added Hc-Tri loss and we see that we get the best performance (row 10) when we have all the four components of CE, Margin MMD-ID, Hc-Tri, and RE augmentation.

\begin{table}[ht]
\begin{center}
\begin{adjustbox}{width=0.7\linewidth}
\centering
\begin{tabular}{ccccccc|cc|cc}
\hline
 \multicolumn{7}{c|}{\textbf{Components}}& \multicolumn{2}{c|}{SYSU-MM01} & \multicolumn{2}{c}{RegDB}\\
 \cline{1-11}
Sr.No & C.E. & HC-Tri & MMD & MMD-ID & Margin MMD-ID & R.E. &r1&mAP&r1&mAP\\
\hline

1&\cmark & \xmark & \xmark & \xmark & \xmark & \xmark & 52.78 & 50.29 & 69.45 (72.94) & 66.31 (69.53)  \\
2&\cmark & \xmark & \cmark & \xmark & \xmark & \xmark & 59.09 & 54.85 & 82.95 (84.66) & 78.63 (80.17)  \\
3&\cmark & \xmark & \xmark & \cmark & \xmark & \xmark & 57.07 & 53.52  & 90.52 (91.02) & 85.59 (86.74) \\
4&\cmark & \xmark & \xmark & \xmark & \cmark & \xmark & 60.13 & 55.97 & 90.76 (91.33)  & 85.31 (85.51) \\
5&\cmark & \xmark & \xmark & \xmark & \cmark & \cmark & \textbf{64.86} & \textbf{60.12} & \textbf{93.57 (93.95)} & \textbf{86.54(88.74)} \\
\midrule
6&\cmark & \cmark & \xmark & \xmark & \xmark & \xmark & 54.75 & 52.14 & 86.18 (88.79) & 80.80 (81.81) \\
7&\cmark & \cmark & \cmark & \xmark & \xmark & \xmark & 59.25 & 55.32 & 89.94 (91.52) & 84.70 (85.92) \\
8&\cmark & \cmark & \xmark & \cmark & \xmark & \xmark & 62.15 & 57.58 & 90.85 (92.68) & 86.53 (87.68) \\
9&\cmark & \cmark & \xmark & \xmark & \cmark & \xmark & 63.11 & 58.48 & 92.44 (93.78) & 87.76 (88.82) \\
10&\cmark & \cmark & \xmark & \xmark & \cmark & \cmark & \textbf{66.75} & \textbf{62.25} & \textbf{93.65 (95.06)} & \textbf{87.30 (88.95)} \\
\bottomrule

\end{tabular}
\end{adjustbox}
\end{center}
\vspace{-5mm}
\caption{Ablation Study of different Components on SYSU-MM01 on RegDB datasets. For RegDB dataset, metrics reported as : Thermal to Visible (Visible to Thermal)}
\label{table:3}
\end{table}

\textbf{Using Margin MMD-ID with existing baselines:}
To further evaluate the generalisability of our Margin MMD-ID, we take three popular and open-sourced baselines: AGW (\cite{ye2021deep}), DGTL (\cite{liu2021strong}) and HcTri \cite{liu2020parameter}. The top-row for each baseline in Table-\ref{table:4} corresponds to the metrics reported in their original work on the SYSU-MM01 dataset. We progressively add MMD-ID and Margin MMD-ID to evaluate their effects on the overall performance. Two goals of this experiment are we want the Margin MMD-ID to be easily integrated with existing baselines without many changes and to get an overall improvement by adding Margin MMD-ID loss in training. It is worth noting that adding Margin MMD-ID loss is not only compatible with the three baselines, but we also get a considerable improvement over baseline (top-row) as well as standard conditional MMD-ID (middle-row). For further details regarding each baseline experiment, please refer to the supplementary material.

\begin{table}[ht]
\begin{center}
\begin{adjustbox}{width=0.7\linewidth}
\centering
\begin{tabular}{c|cccc}
\hline
 \textbf{Method} & \multicolumn{4}{c}{SYSU-MM01} \\
 \cline{2-5}
 & r1 & r10 & r20 & mAP\\
\hline
AGW & 47.50 (54.17) & 84.39 (91.14) &  92.14 (95.98) & 47.65 (62.97) \\
AGW + MMD-ID & 53.10 (58.05) & 89.97 (96.03) & 95.83 \textbf{(99.32)} & 51.12 (66.41) \\
AGW + Margin MMD-ID & \textbf{54.35 (59.17)}  & \textbf{90.87 (96.09)} & \textbf{96.09} (99.27) & \textbf{51.91 (66.92)} \\
\midrule
DGTL & 57.34 (63.11) & - & - & 55.13 (69.20) \\
DGTL + MMD-ID & 58.77 (62.75) & 90.94 (94.96) & 96.01 (98.73) & 55.59 (68.99) \\
DGTL + Margin MMD-ID & \textbf{59.63 (65.13)}  & \textbf{92.10 (96.17)}  & \textbf{96.84 (99.15)}  & \textbf{56.50 (71.26)}  \\
\midrule
HcTri & 61.68 (63.41) & \textbf{93.10} (91.69) & \textbf{97.17} (95.28) & 57.51 (68.17) \\
HcTri + MMD-ID & 63.50 (67.18) &  92.11 (93.32) & 96.47 (97.14) & 59.69 (71.81) \\
HcTri + Margin MMD-ID & \textbf{64.35 (68.49)}   &  93.02 \textbf{(93.55)} & 96.96 \textbf{(97.33)} & \textbf{60.11 (72.73)} \\ 
\bottomrule

\end{tabular}
\end{adjustbox}
\end{center}
\vspace{-4mm}
\caption{Incorporating Margin MMD-ID on existing baselines (AGW \cite{ye2021deep}, DGTL \cite{liu2021strong}, HcTri \cite{liu2020parameter}) for SYSU-MM01 dataset. For each setting metrics are reported as: All-Search (Indoor-Search)}
\label{table:4}
\end{table}

\begin{figure}[h]
\centering
\includegraphics[width=0.9\textwidth,height=3cm]{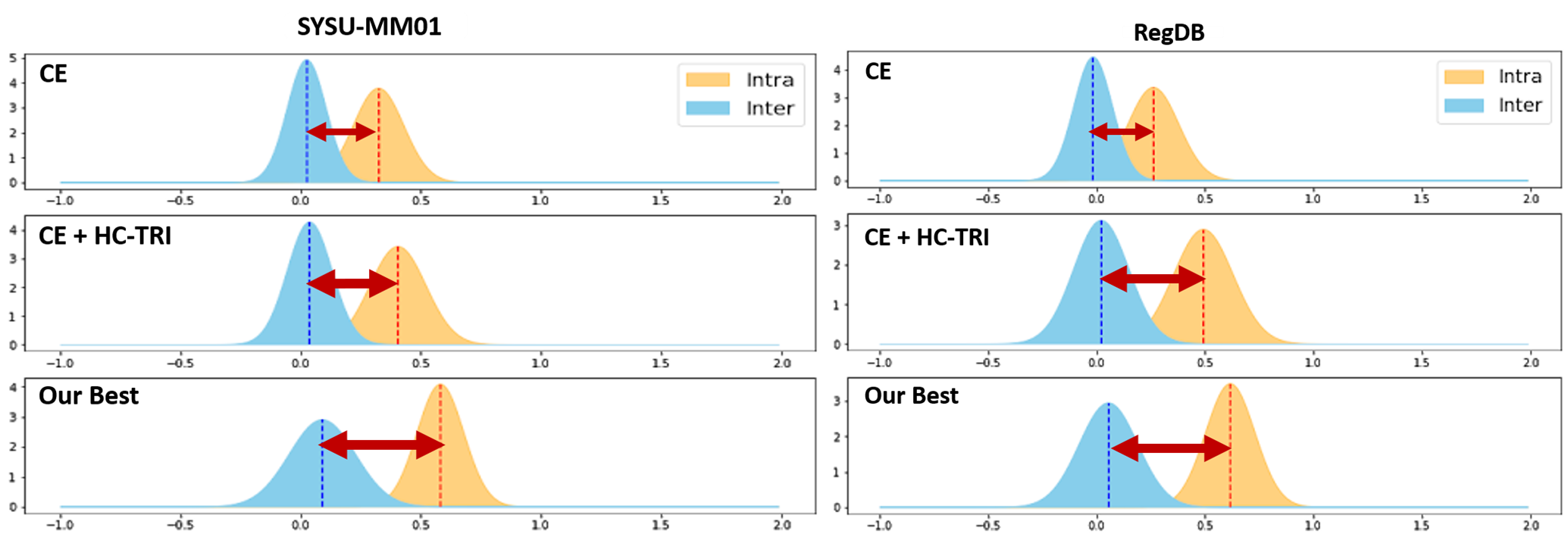}
\caption{Plot for Gaussian fitted distributions over given mean(m) and std deviation(s) for Intra and Inter class similarities on Test identities.}
\label{fig:Gaussian}
\end{figure}

\textbf{Qualitative evaluation: }
To visualize the inter-class separation and intra-class compactness across the modalities (shown in Fig.\ref{fig:Gaussian}), we define a thermal and visible feature representative for each identity by calculating the centroid of image features belonging to that identity and modality. Thus, we have a visible and thermal feature vector for each identity.
Ideally, discriminative yet modality-invariant features should give high intra-class and low inter-class similarity values. We calculate the intra-class similarity by finding the cosine distance between each identity's visible and thermal centroid features and calculate the mean and standard deviation, on which we fit a Gaussian distribution (Orange curve in Fig.\ref{fig:Gaussian}). Similarly, we calculate the inter-class similarity by finding the cosine distances between the visible and thermal centroid features of different identities and get the mean and standard deviation and fit a Gaussian distribution (Blue curve). Fig.\ref{fig:Gaussian} shows that the intra-class similarity between visible and thermal pairs has increased, indicating the feature vectors of different modalities for same identity are more closer when we use Margin MMD-ID loss. As a result, the separation between the inter and intra class similarities has increased, which is needed to avoid misclassifications. To avoid outliers, we use centroids for each identity instead of individual samples. We choose this strategy of using all identities (then fitting a Gaussian over mean and standard deviation), instead of selecting few identities, so as to holistically visualise the inter-class and intra-class similarities.

\textbf{Ablation study for Margin: }
We find the optimal margin value by following the similar strategy as employed by conventional methods \cite{WangDualDisc}, \cite{Dai2018CrossModalityPR}, \cite{liu2021strong} i.e., using validation data to tune the hyperparameters. Specifically, since ‘$\rho$’ is a hyper-parameter, we fine-tune it separately on both datasets. We perform a sensitivity analysis for the margin values (Fig.\ref{Figure:4} for RegDB and Supplementary Fig.3 for SYSU-MM01), which conveys that the performance is stable across a broad range of margins ($\rho$), around the optimal. Consequently, we choose $\rho$ as 1.4 for both SYSU-MM01 and RegDB as it’s the best performing margin for both datasets. It is worth noting that the stable nature of Margin MMD-ID for our configuration allowed us to keep same margin across both datasets.


\textbf{Computational cost analysis: }
We create our batch with $2\times PK$ images, where $P$ is number of identities and $K$ is number of visible and thermal images. $L_{Margin-MMD-ID}$ requires computing $PK(K-1)$ pairwise distances for same distribution term and $P \times K \times K$ distances for cross distribution term, which after summing makes a total of $PK[2K-1]$ computations. This is comparable with the computations needed for a batch with $2\times PK$ images for a standard triplet loss \cite{liu2020parameter} which is $2PK(2K−1)$ for hardest positive sample mining and $2PK\times 2(P−1)K$ for hardest negative sample mining. Also, Hc-Tri loss \cite{liu2020parameter}, for a batch requires $P$ computations for positive and $2P\times 2(P-1)$ for negative term. Thus, computation wise, our loss is comparable to standard Triplet loss. We also do a training time analysis and observe the hours needed to train a model for 60 epochs for different setups. Table \ref{Table:5} shows that using MMD-ReID negligibly increases the training time over C.E. and C.E. + HC-Tri.

\begin{figure}[h]
\begin{floatrow}
\ffigbox{%
    \includegraphics[width=0.5\textwidth]{images/Final_Plot_regdb.pdf}%
}{%
  \caption{Sensitivity analysis for Margin on RegDB}%
  \vspace{-3mm}
  \label{Figure:4}
}
\capbtabbox{%
  \begin{tabular}{c|c} \hline
    Setup & Training time \\ \hline
    C.E. & 5.45 hrs \\
    C.E. + HC-Tri & 5.81 hrs \\
    MMD-ReID (All loss)  & 6 hrs \\
    \hline
  \end{tabular}
}{%
  \caption{Training time analysis on Nvidia GTX 1080Ti: SYSU-MM01}%
  \vspace{-3mm}
  \label{Table:5}%
}
\end{floatrow}
\end{figure}


\vspace{-4mm}
\section{Conclusion}
\label{sec:conclusion}

Although the last few years have witnessed significant progress in the VT-ReID task, the current state-of-the-art methods aim to reduce the cross-modality discrepancy in an implicit fashion by aligning pixel and feature space representations using adversarial learning strategies or designing domain-knowledge reliant feature extraction modules. This paper provides a simple but effective framework for performing VT-ReID called MMD-ReID based on a margin-modification of the standard MMD. We empirically observed that using standard MMD to align identity-conditioned visible and thermal distributions in supervised VT-ReID task leads to overfitting and devise a simple margin-modification, Margin MMD-ID, to alleviate it. Extensive experimentations demonstrate the superiority of our proposed framework as well as validate the effectiveness of each component in it. We also evaluate the effect of incorporating Margin MMD-ID in existing baselines and observe that it leads to significant gains in performance. We thus urge the VT-ReID community to explore more simpler and stronger ways to solve this problem of VT-ReID.

\bibliography{egbib}
\end{document}


\maketitle
\section{Architecture and Loss details}
\textbf{GEM pooling layer:}
To get fine grained features, our two stream network is terminated by a Generalised Mean Pooling layer \cite{radenovic2018fine, ye2021deep}, which is defined as: 
\begin{equation}\label{eq:9}
f=[f_1, f_2,... f_K]^T, f_k = \frac{1}{\mid X_k \mid}(\sum_{x \in X_k} x^{p_k} )^{1/p_k}
\end{equation}
where \(f_k\) is a feature map, K is the number of feature maps input to GeM pooling, \(X_k\) is the set of pixels in a HxW shaped feature activation map say. The output of GeM layer is a 1-D vector with each component representing one feature map.
\newline
\newline
\textbf{HC-Tri loss:}
Triplet loss \cite{hermans2017defense} is a widely used metric learning loss in Person ReID. Each mini-batch sample is considered as an anchor, and the hardest positive and hardest negative sample is selected for this anchor. To effectively fetch positives in the mini-batch, the mini-batch is formed by randomly sampling P identiies and randomly sampling K images of each identity, resulting in a mini-batch with PK images. This loss compares each sample (anchor) to all other samples which is a strict constraint, perhaps too strict to constrain the pairwise distance if there exist some outliers (bad examples), which would form the adverse triplet to destroy other pairwise distances \cite{liu2020parameter}. Therefore, \cite{liu2020parameter} considers adopting the center of each person as the identity agent. In this manner, we can relax the strict constraint by replacing the comparison of the anchor to all the other samples by the anchor centre to all the other centres.

$\begin{aligned}
L_{h c_{-} t r i}(C)=& \sum_{i=1}^{P}\left[\rho+\left\|c_{v}^{i}-c_{t}^{i}\right\|_{2}-\min _{n \in\{v, t\}, j \neq i}\left\|c_{v}^{i}-c_{n}^{j}\right\|_{2}\right]_{+} \\
&+\sum_{i=1}^{P}\left[\rho+\left\|c_{t}^{i}-c_{v}^{i}\right\|_{2}-\min _{n \in\{v, t\}, j \neq i}\left\|c_{t}^{i}-c_{n}^{j}\right\|_{2}\right]_{+}
\end{aligned}$

where, 
$\begin{aligned}
\; \; c_{v}^{i} &=\frac{1}{K} \sum_{j=1}^{K} v_{j}^{i},  \; \;\;
 \; \;\; c_{t}^{i}&=\frac{1}{K}\sum_{j=1}^{K} t_{j}^{i}
\end{aligned}$

$\{c_v^i |i=1,2,...P \}$ are the visible centres and $\{c_t^i |i=1,2,...P \}$ are the thermal centres. $L_{h c_{-} t r i}$ concentrates on only one cross-modality positive pair and the mined hardest negative pair in both the intra and inter-modality.

\section{Implementation details}

\subsection{MMD-ReID:}
We adopt ResNet50 \cite{he2016deep} as the backbone network. The stride of the last convolution layer is changed from 2 to 1 to get fine-grained features \cite{sun2018beyond}. Input images are resized to 288x144 shape and padded with 10, followed by Data augmentation techniques like random cropping of 288x144 shape and Random Horizontal flipping. We also use Random erasing augmentation \cite{zhong2020random} with probability 0.5 for some experiments, which we discuss in the Results section of the main paper. We use a Stochastic Gradient descent optimizer (SGD) with momentum as 0.9 and 0.0005 weight decay. We set initial lr as 0.01 for ResNet50 parameters and 0.1 for BatchNorm layer and Classifier (FC layer) for both datasets (SYSU-MM01 and RegDB). Warmup learning rate strategy is applied to improve performance as \cite{liu2020parameter}. For sampling, we choose P and K both as 4 for both datasets. Margin \(\rho\) for Margin MMD-ID loss is set as 1.4 for both the datasets and \(\rho_1\) for HC-Tri loss is 0.3. The tradeoff parameters in total loss equation of main paper: \(\lambda_1, \lambda_2, \lambda_3\) are set as 1, 0.25,2. We train our model on a single Nvidia GTX 1080Ti gpu card for 60 epochs which takes $\sim$ 6 hours to train for SYSU-MM01 and $\sim$ 1.3 hours for RegDB with all our losses.

\section{Ablation Study}

\subsection{Effect of Random erasing augmentation}
Random Erasing (RE) augmentation \cite{zhong2020random} is a well-known regularisation technique that helps in improving the generalisation ability of the model. We incorporate RE with our total loss formulation to get better performance. To ensure that the gain in performance is not because of adding RE, we perform a set of experiments with RE and without RE to see the net effect of adding RE. Table \ref{table:1} shows the experiments with the corresponding rank-1 and mAP values. It is evident from the last two rows that even without adding RE, our final Margin MMD-ID loss along with Cross entropy and HC-Tri loss (row 7) performs comparably with the state of the art models. Adding RE (row 8) gives the boost hence we use RE in our final model. Also, adding RE with only Cross entropy loss (row 2) or with Cross entropy and HC-Tri loss (row 4) doesn't give much performance boost as RE on itself, cannot reduce the modality gap.

\begin{table}[h]
\begin{center}
\begin{adjustbox}{width=0.7\linewidth}
\centering
\begin{tabular}{cl|cc}
\hline

Sr. No & Method & r1 &mAP\\
\midrule

1& C.E. & 52.78 & 50.29 \\
2& C.E. (w R.E.) & 55.32 & 51.24 \\
\midrule

3& C.E. + HC-Tri & 54.75 & 52.14 \\
4& C.E. + HC-Tri (w R.E.) & 60.94 & 55.39 \\
\midrule

5& C.E. + HC-Tri + MMD-ID & 62.15 & 57.58 \\
6& C.E. + HC-Tri + MMD-ID (w R.E.) & 64.4 & 59.8 \\
\midrule

7& C.E. + HC-Tri + Margin MMD-ID & 63.11 & 58.48 \\
8& C.E. + HC-Tri + Margin MMD-ID (w R.E.) & \textbf{66.75} & \textbf{62.25} \\
\midrule

\end{tabular}
\end{adjustbox}
\end{center}
\caption{Effect of Random Erasing (R.E.) augmentation on different components in MMD-ReID. Results provided for All-Search mode in SYSU-MM01 dataset}
\label{table:1}
\end{table}



\subsection{Dataset Complexity: RegDB}
\begin{figure}[ht]
  \centering
  \includegraphics[width=\linewidth]{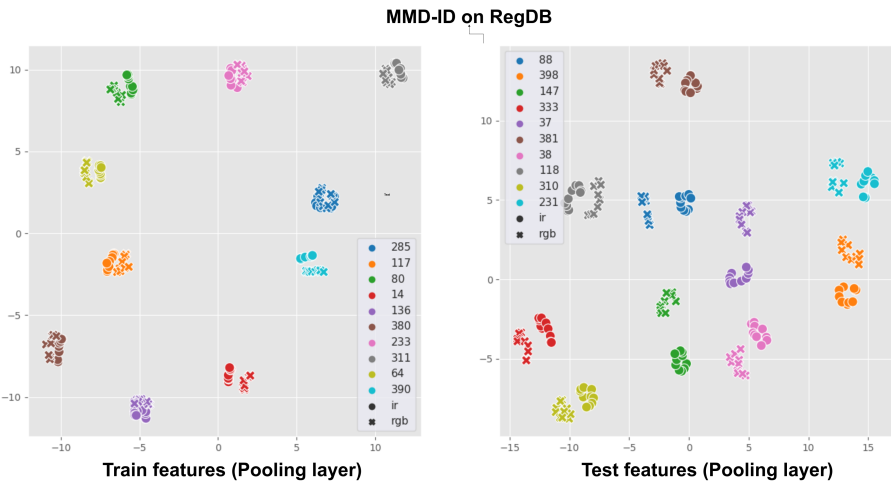}
  \caption{t-SNE visualisation on RegDB which shows the features are easily separable and less prone to overfitting.}
  \label{fig:Regdb_nooverfit}
\end{figure}

RegDB \cite{nguyen2017person} is collected from two well-aligned cameras (one visible and one thermal), compared to six cameras for SYSU-MM01 (four visible and two thermal in both indoor and outdoor environments). For RegDB evaluation, the dataset is randomly split into two parts, one for training and one for testing. Thus, for each modality (e.g., visible), the samples during training and testing are captured using the same camera. This eliminates significant intra-modality variations (such as viewpoint and pose changes), usually caused when images are captured using multiple cameras. Moreover, SYSU-MM01 (38,271) has more than four times the number of samples present in the RegDB dataset (8,240), further increasing the complexity of matching identities across modalities. The aforementioned reasons indicate that RegDB is a much simpler dataset to operate on with less vulnerability to overfitting due to train-test sampling similarities. Thus, applying MMD-ID on RegDB doesn’t correspond to feature-degradation or overfitting and provides relatively decent performance compared to evaluation on SYSU-MM01 (Table-3 in the main paper, row-2;3). We also empirically verify this insight by generating the t-SNE plots for MMD-ID on the RegDB dataset. We observe that both train and test features demonstrate high inter-class separation and intra-class compactness (Fig.\ref{fig:Regdb_nooverfit}). Fig.\ref{fig:Regdb_nooverfit} reveals that the features for each identity are easily separable and consequently have little chance of overfitting. Lastly, recent state-of-the-art works \cite{liu2020parameter} have also observed a similar high performance on the RegDB dataset (compared to SYSU-MM01). 

\subsection{Qualitative visualisation using T-SNE}
\begin{figure}[ht]
  \centering
  \includegraphics[width=\linewidth]{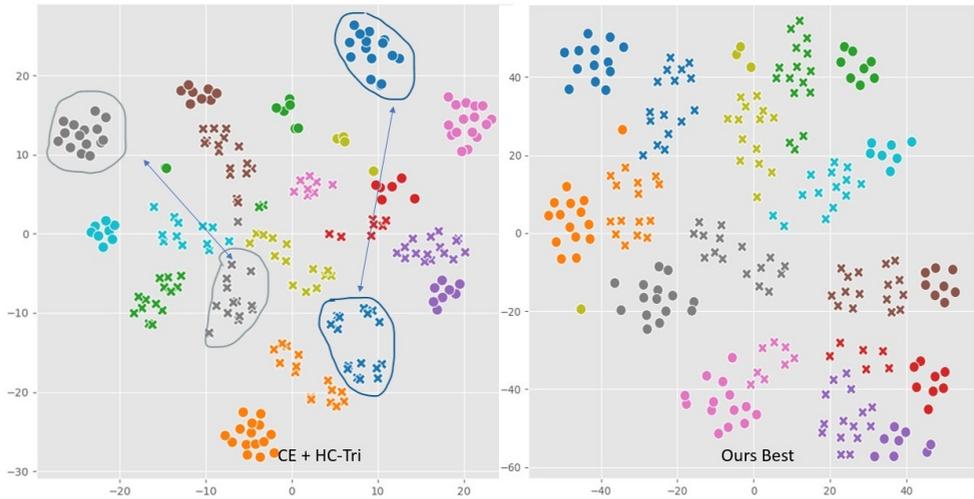}
  \caption{T-SNE visualisation on ten randomly sampled test identities (SYSU-MM01) for CE+HC-Tri loss trained model (baseline) Vs Our Best (MMD-ReID) model. Different color denotes different identities. Cross and circle marker denotes thermal and visible features respectively.}
  \label{fig:tsne_sysu}
\end{figure}

Figure \ref{fig:tsne_sysu} shows the qualitative visualisation of the features after the BatchNorm layer, using T-SNE plots \cite{van2008visualizing}. The left side plot is for the features (belonging to test-data) extracted by a model trained with only Cross entropy (CE) + HC-Tri loss, and the Right side plot is for the model trained with our MMD ReID framework. It is clear from the Left side plot that the visible and thermal features for a particular identity form separate clusters and are well separated, which is undesirable. The visible and thermal clusters ideally should be compact and as close as possible to avoid misclassifications. The right side plot has successfully achieved these properties by bringing the same identity visible and thermal features closer in feature space. Thus, the visual analysis also supports our MMD-ReID framework.

\subsection{Implementation details for Margin MMD-ID with existing baselines:}
To verify generalisation capability of MMD-ReID, we take three popular and open-sourced baselines and add MMD-ID and Margin MMD-ID losses on them. The details about the baselines and hyperparameters used are described below. Note that the table for accuracies with MMD on different baselines is presented in main paper, Table 4.
\newline

\textbf{AGW} (Average Generalized mean pooling with Weighted triplet loss): Ye et al. in their work \cite{ye2021deep} introduced a new powerful baseline for Person Re-ID. AGW proposed three major modifications on top of the best practices discussed in \cite{Luo2019BagOT}: Non-local attention blocks, Generalized-mean (GeM) pooling layer, and Weighted regularized triplet loss. In line with the standard setup, the MMD-ID and Margin MMD-ID losses are computed on features extracted from the GeM layer while features extracted from the BatchNorm layer are used during inference time. The margin (\(\rho\)) in Margin MMD-ID is set as 0.4 whereas all other hyperparameters are kept the same as reported by \cite{ye2021deep}.
\newline

\textbf{DGTL }(Dual-Granularity Triplet Loss): DGTL \cite{liu2021strong} utilizes sample-based and center-based triplet loss in a hierarchical manner to encourage intra-class compactness and inter-class discrimination at fine and coarse granularity levels simultaneously. This setup allows achieving competitive performance without the need for aggregating local-level features via architectural improvements. In accordance with previous experiments, we employ the MMD-ID and Margin MMD-ID loss on the features extracted from the pooling layer (in the fine granularity level branch). The margin (\(\rho\)) in Margin MMD-ID is set as 1.00 while all other hyperparameters remain unchanged.
\newline

\textbf{HcTri }(Hetero-center Triplet Loss): Since traditional triplet-loss is prone to outliers and often fails to converge, Liu et al. \cite{liu2020parameter} in their work proposed a novel hetero-center triplet loss that operates on a coarse granularity level. The Hc-Tri loss in a part-based person feature learning framework leads to superior performance than the standard triplet loss. The Hc-Tri loss is computed for each part-level feature strip as well as the final concatenated global features. For a fair comparison with other baselines,  we employ the MMD-ID and Margin MMD-ID loss only on the concatenated global feature vector. The margin (\(\rho\)) in Margin MMD-ID is set as 1.00 while all other hyperparameters are kept the same. 

\subsection{Ablation study for Margin}
\begin{figure}[ht]
  \centering
  \includegraphics[width=\linewidth]{images/margin_plot.png}
  \caption{Line plot to observe the effect of margin variations on the performance of model in terms of rank-1 and mAP values. Results are obtained on SYSU-MM01 dataset for All search single shot mode.}
  \label{fig:Figure 3}
\end{figure}

In continuation of the ‘Ablation study for Margin’ result on the RegDB dataset presented in section 4.2 of the main paper, here we discuss the same for the SYSU-MM01 dataset. To study the stability of our model with varying margin values, we change the value of $\rho$ in a set of \{0.8, 1.0, 1.2, 1.4, 1.6, 1.8\} and observe the rank-1 and mAP values with our total loss formulation. The line plot for the same is shown in Fig.\ref{fig:Figure 3}. We can see that $\rho$ = 1.4 gives the best results.

\bibliography{egbib}